# GDA-YOLO11: Amodal Instance Segmentation for Occlusion-Robust Robotic Fruit Harvesting

Caner Beldek, Emre Sariyildiz, Son Lam Phung, and Gursel Alici

*Abstract*—Occlusion remains a critical challenge in robotic fruit harvesting, as undetected or inaccurately localised fruits often results in substantial crop losses. To mitigate this issue, we propose a harvesting framework using a new amodal segmentation model, GDA-YOLO11, which incorporates architectural improvements and an updated asymmetric mask loss. The proposed model is trained on a modified version of a public citrus dataset and evaluated on both the base dataset and occlusion-sensitive subsets with varying occlusion levels. Within the framework, full fruit masks, including invisible regions, are inferred by GDA-YOLO11, and picking points are subsequently estimated using the Euclidean distance transform. These points are then projected into 3D coordinates for robotic harvesting execution. Experiments were conducted using real citrus fruits in a controlled environment simulating occlusion scenarios. Notably, to the best of our knowledge, this study provides the first practical demonstration of amodal instance segmentation in robotic fruit harvesting. GDA-YOLO11 achieves a precision of 0.844, recall of 0.846, mAP@50 of 0.914, and mAP@50:95 of 0.636, outperforming YOLO11n by 5.1%, 1.3%, and 1.0% in precision, mAP@50, and mAP@50:95, respectively. The framework attains harvesting success rates of 92.59%, 85.18%, 48.14%, and 22.22% at zero to high occlusion levels, improving success by 3.5% under medium and high occlusion. These findings demonstrate that GDA-YOLO11 enhances occlusion-robust segmentation and streamlines perception-to-action integration, paving the way for more reliable autonomous systems in agriculture.

*Index Terms*—Amodal instance segmentation; robotic harvesting; Occlusion; YOLO11; Agricultural robotics.

## I. INTRODUCTION

UNSUSTAINABLE agricultural practices contribute to significant food waste, directly undermining the economic stability of farmlands, exacerbating environmental degradation, increasing greenhouse gas emissions, and threatening global food security [1–3]. Among the various factors, conventional inefficient harvesting techniques stand out as a primary cause of food waste, resulting in substantial crop losses and resource wastage throughout the food production chain [4]. As the world faces mounting challenges such as hunger, increasing food demand and resource scarcity, there is an urgent need to modernise our harvesting practices to enhance productivity and ensure sustainability [1]. Recently, robotic technology has gained increasing importance in sustainable agricultural harvesting [5–7], primarily due to intelligent harvesters equipped with various sensors and actuators. These systems are capable of gently picking a wide range of crops without human intervention [7].

One of the fundamental requirements for such robotic harvesting systems is the ability to perceive their environment. In this context, target fruit recognition plays a critical role, as it marks the first step of the harvesting process, and the entire pipeline depends on the accurate localisation of the identified fruit. In the literature, fruit recognition tasks are commonly addressed by leveraging computer vision (CV) strategies in conjunction with deep learning (DL) [6]. This trend is driven by the widespread use of Red-Green-Blue-Depth (RGB-D) cameras in practical applications, due to their affordable cost and 3D sensing capabilities [8].

Fruit recognition in agricultural settings has been widely addressed using CV tasks, including object detection [9], semantic segmentation [10], and instance segmentation [11–13], typically implemented using DL models. Object detection methods in agricultural robotics estimate fruit locations using bounding boxes, offering fast but coarse localisation. In contrast, segmentation methods generate precise masks that enable the pixel-level accuracy required for robotic harvesting. Semantic segmentation typically classifies each pixel, but cannot distinguish between individual fruits, limiting its use in precise harvesting tasks that need spatial features such as location, orientation, and size. Consequently, instance segmentation offers a more suitable framework for precise harvesting tasks, as it provides object-level differentiation and preserves essential spatial features required for accurate manipulation [12].

Despite growing interest and successful applications in fruit recognition, most existing vision-based DL approaches focus solely on the visible portions of target fruits, offering no solution for occluded regions [14]. However, partial occlusion, which is caused by leaves or surrounding plant organs, is a prevalent issue in agricultural environments. It not only degrades recognition accuracy but also disrupts the localisation of the picking point, which is directly inferred from the prediction. As a result, fruits may remain unrecognised or be recognised inaccurately, leading to incorrect picking points. These problems can ultimately hinder the harvesting, causing missed picks, fruit damage, or mechanical collision [15]. However, occlusion represents not

C. Beldek, E. Sariyildiz, S. L. Phung and G. Alici are with the School of Engineering, University of Wollongong, 2522, NSW, Australia (e-mail: cb018@uowmail.edu.au, emre@uow.edu.au, phung@uow.edu.au, gursel@uow.edu.au).

only a vision-based sensing challenge but also a physical manipulation constraint for the robotic arm and end-effector [16].

To mitigate the occlusion effect in agricultural fields, earlier studies have approached this problem through depth-based shape reconstruction and geometric modelling. For example, Gong et al. [17] proposed a method to reconstruct the full 3D shape of an occluded tomato using fused depth maps, allowing estimation of its centre and radius before grasping. Furthermore, shape-fitting techniques leveraging prior knowledge of fruit geometry, such as spheres or ellipsoids, have been employed to infer the hidden structure of partially visible fruits [18, 19]. However, relying solely on prior shape assumptions limits adaptability in agricultural settings, where scenes are inherently complex, fields are unstructured, and fruit appearance varies greatly, making such assumptions unreliable for consistent identification [14]. Other approaches have focused on occlusion-aware segmentation by incorporating shape priors and occlusion recovery algorithms [20], or by integrating multi-view data and point cloud completion [21]. Nevertheless, these segmentation pipelines frequently involve complex and time-consuming multi-step processes, including visible mask extraction, camera calibration, or geometry-based completion, and generally lack validation in practical robotic harvesting scenarios [22].

Unlike methods that address occlusion by focusing only on the visible regions, amodal instance segmentation aims to predict the complete shape of an object, which is crucial for accurate fruit positioning and precise grasping [23]. Similar to standard instance segmentation, amodal instance segmentation individually detects each fruit and captures spatial features, but it also covers predicting occluded regions. Gené-Mola et al. [23] were among the first to explore this concept in fruit imagery, estimating complete apple sizes under occlusion using an amodal instance segmentation model in 2023. Yang et al. [24] and Kim et al. [14] extended this research to tomatoes and cucumbers, respectively, applying amodal segmentation for shape reconstruction and occlusion recovery. However, these studies did not integrate their results into a full harvesting pipeline. More broadly, related works on dataset development and benchmarking [25] often lack validation through physical experiments or real hardware implementations, limiting practical relevance. While prior studies have explored amodal segmentation for estimating full fruit shape under occlusion, these methods largely remain at the perception level and have not been extended to robotic harvesting. The lack of integration between amodal perception and physical interaction limits their practical utility, as most approaches do not validate whether predicted full-object masks translate into effective robotic actions. Bridging this gap is essential to assess the true relevance of amodal reasoning in field robotics and to advance beyond isolated perception tasks toward actionable, deployment-ready systems.

To this end, this paper presents a perception-to-action harvesting framework based on a novel amodal instance segmentation model, built upon the nano-sized YOLO11 architecture. A Global Attention Module (GAM) was integrated into the model's structure by embedding one at the end of YOLO11's neck, and the Cross Stage Partial with Spatial Attention (C2f-PSA) block was replaced with another GAM. In addition, the kernel size of the Spatial Pyramid Pooling-Fast (SPPF) block, which is at the end of the backbone, was increased to enhance the receptive field. Furthermore, the baseline's head was deepened, and its standard mask loss function was updated with an asymmetric mask loss. This new loss was specifically designed to handle occlusion by penalising false negatives more heavily than false positives, thereby encouraging the model to preserve partial object masks. The developed model yielded 84.4% precision, 84.6% recall, 91.4% mAP@50, and 63.6% mAP@50:95, outperforming YOLO11 by 5.1% in precision, 1.3% in mAP@50 and 1% in mAP@50:95, while increasing the number of parameters by only 18%. To thoroughly validate the enhanced model, predictions on the occlusion-specific subsets and robotic harvesting experiments were conducted. Under occlusion, harvesting scenarios were run in a controlled lab environment with a real fruit setup. The harvesting outcomes demonstrated that GDA-YOLO11 achieved picking performances of 92.59% without occlusion, 85.18% under low occlusion, 48.14% under medium occlusion, and 22.22% under high occlusion. These results confirm that the proposed improvements surpassed the baseline model by over 3.5% in challenging medium and high occlusion scenarios. Additionally, a strong correlation between mAP@50 scores of GDA-YOLO11's occlusion-sensitive subsets and harvesting performance was observed across all corresponding occlusion classes, evidenced by an $R^2$ of approximately 0.986. This strong alignment between segmentation accuracy and harvesting success not only underscores the practical value of amodal perception but also establishes its viability as a core enabler for reliable, occlusion-robust operation in practical agricultural harvesting robotics systems.

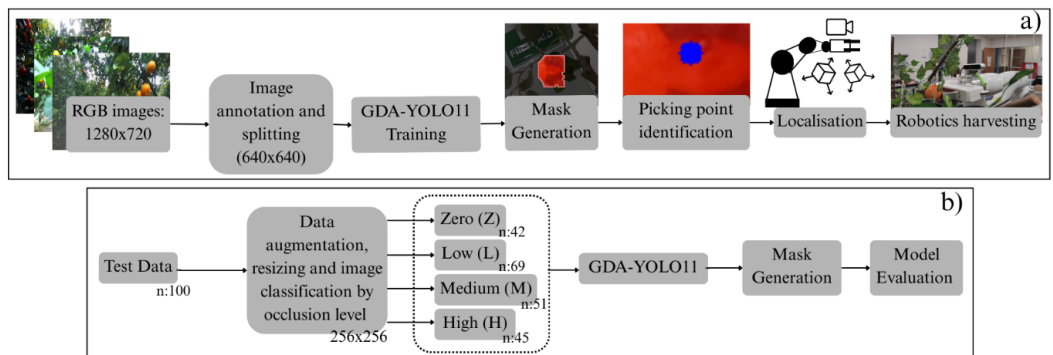

Fig. 1. Overview of the proposed framework. **(a)** The process workflow, ranging from raw data acquisition to the harvesting application. **(b)** The technical pipeline for occlusion-sensitive tests, including, augmentation, resizing, model training, and mask generation.

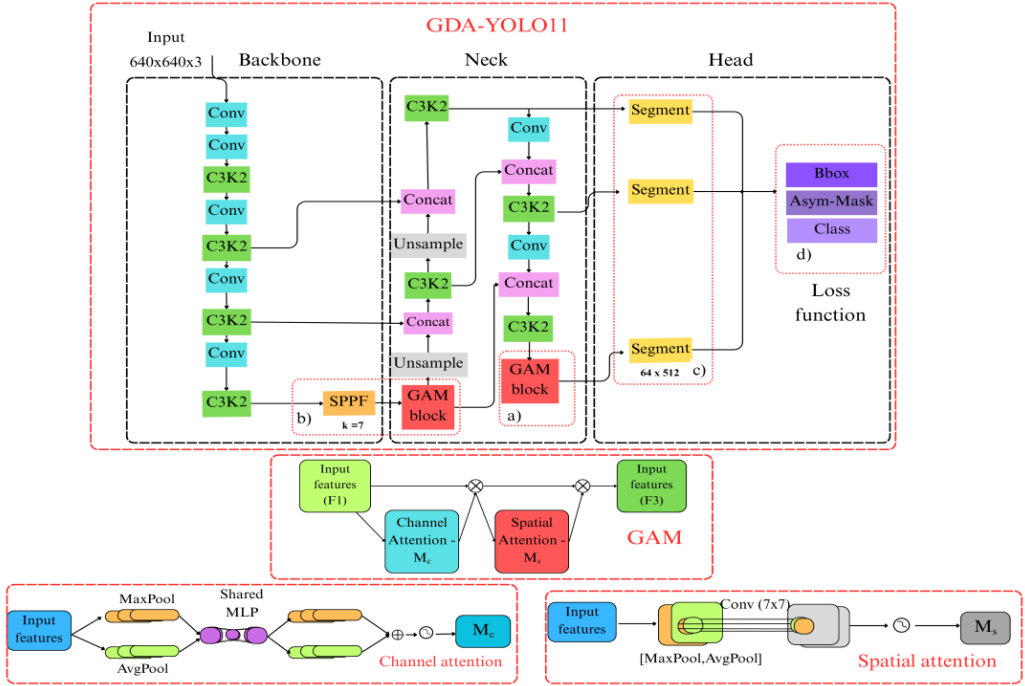

Fig. 2. Overall architecture of GDA-YOLO11. a) GAM block, b) GAM block replacing the C2F-PSA, and kernel extension of SPFF, c) deepened segmentation head, and d) updated mask loss function with asym-mask.

Within this context, the main contributions of this paper are as follows: 1) the development of GDA-YOLO11, a new amodal instance segmentation model that extends the lightweight YOLO11 architecture with explicit occlusion handling capabilities, specifically designed to address partial occlusions in fruit harvesting scenarios; and 2) the integration of amodal perception into a practical robotic harvesting application (Fig. 1), marking, the first known demonstration of amodal instance segmentation applied to a controlled fruit harvesting application.

The rest of the paper is organised as follows. Section II defines the methodology to introduce the proposed novel model structure with improvements. Section III presents the training strategy, including the dataset preparation, and evaluation metrics, Section IV introduces harvesting framework with all subworks. Section V includes results and discussion in terms of the proposed model and fruit picking framework. Finally, Section VI concludes the paper.

## II. METHODOLOGY

### A. Architecture overview

The GDA-YOLOv11 retains the standard YOLO11 backbone, where features pass through only one C3k2 block before multi-scale fusion in the neck. As a single-stage, anchor-free detector, YOLOv11 optimises feature extraction and spatial representation through integrated attention mechanisms. While the baseline typically employs a standard, lightweight segmentation head and a standard Binary Cross Entropy (BCE) mask loss, these core components were modified to enhance amodal segmentation under occlusion, as illustrated in Fig. 2. First, the network's neck was modified to create a dual-attention structure by adding a Global Attention Module (GAM) at its end while also replacing the initial C2f-PSA block with a second GAM as shown in Fig. 2a and 2b. Concurrently, the SPPF kernel size was expanded to 7x7 to provide a wider receptive field for superior spatial context aggregation. Second, the segmentation head was structurally deepened to better resolve the fine-grained boundaries of partially visible objects (Fig. 2c). Finally, an asymmetric mask loss function was implemented, using a weighted Binary Cross Entropy (BCE) that more heavily penalizes false negatives to improve recall in cluttered scenes as shown in Fig. 2d. The proposed architecture is named GDA-YOLOv11 to reflect these three core enhancements (GAM integration, a Deepened head, and an Asymmetric mask loss) and is optimized for robust performance in complex and occluded visual environments.

### B. GAM

As illustrated in Fig. 2, the GAM [26] is an attention mechanism that enhances both channel-wise and spatial feature representations by sequentially applying channel attention followed by spatial attention within a residual framework. These attentions are applied in a multiplicative manner to refine the features and compute the intermediate state and output feature map as described in Eqs. (1, 2).

$$F_2 = (\mathcal{M}_C(F_1) \otimes (F_1)) \quad (1)$$

$$F_3 = (\mathcal{M}_S(F_2) \otimes (F_2)) \quad (2)$$

where $F_1 \in \mathbb{R}^{C \times H \times W}$ is the input tensor, with $C$, $H$, and $W$ denoting the number of channels, height, and width, respectively; $F_2$ is the intermediate state feature map; $F_3$ is the output map; $\mathcal{M}_C$ is the channel attention map; $\mathcal{M}_S$ is the spatial attention map; and $\otimes$ denotes element-wise multiplication. This structure allows the module to learn both what to focus on (channel-wise) and where to focus (spatially), enhancing semantic consistency and improving the network's ability to identify occluded or ambiguous object regions. The sequential design ensures that global and local cues are captured hierarchically, while the residual path facilitates stable gradient flow during training.

The channel attention module computes a weight vector highlighting important feature channels by aggregating spatial information using average pooling $z_{avg} = \text{AvgPool}(F_1)$ and max pooling $z_{max} = \text{MaxPool}(F_1)$ as defined in (3).

$$z = \sigma(\text{MLP}(z_{avg}) + (z_{max})) \quad (3)$$

where $\sigma$ is the sigmoid activation, MLP denotes a shared multi-layer perceptron consisting of two convolutional layers reducing and restoring the channel dimension with a reduction ratio equal to 16. The output channel attention map $\mathcal{M}_C(F_1)$ is broadcasted spatially and multiplied element-wise with the input $F_1$.

The spatial attention module focuses on highlighting important spatial regions by compressing the channel dimension, employing average pooling $f_{avg} = \text{AvgPool}(F_2) \in \mathbb{R}^{1 \times H \times W}$ and max pooling $f_{max} = \text{MaxPool}(F_2) \in \mathbb{R}^{1 \times H \times W}$ feature maps, described as:

$$\mathcal{M}_S(X) = \sigma(f^{7 \times 7}(D)) \quad (4)$$

where $\sigma$ denotes sigmoid activation; $f^{7 \times 7}$ is a convolutional layer with a 7×7 kernel to capture spatial relationships across a larger context; and D is the concentration of average-pooled and max-pooled spatial summaries, which is Concat[AvgPool(X), MaxPool(X)]. Finally, this attention map is applied element-wise to each channel of the input feature map. The combination of these two modules enables the network to focus selectively both across and within feature maps, making GAM particularly effective in handling occluded scenes.

*C. Deep Head*

To enhance the segmentation performance under occlusion, the original YOLO11n prediction head was replaced with a deeper structure. Specifically, the number of intermediate feature channels was increased from 32 to 64, and the input feature dimension to the final segmentation block was expanded from 256 to 512. This modification allows the model to extract richer feature representations before prediction, leading to more accurate segmentation in visually complex or overlapping object scenarios. While the prediction output dimensions remain unchanged, the deeper feature fusion improves the model's ability to resolve ambiguity and preserve object boundaries under challenging conditions

*D. Asymmetric Mask Loss Function*

The proposed model adopts a hybrid loss function composed of three components: bounding box regression, asymmetric mask segmentation, and classification. These components are jointly optimised to improve detection and instance segmentation performance under occlusion. The total loss is formulated as:

$$\mathcal{L}_{total} = \lambda_{box} \cdot \mathcal{L}_{CIoU} + \lambda_{mask} \cdot \mathcal{L}_{asym-BCE} + \lambda_{cls} \cdot \mathcal{L}_{BCE} \quad (5)$$

where $\mathcal{L}_{CIoU}$ denotes the CIoU loss for bounding box regression; $\mathcal{L}_{asym-BCE}$ is the asymmetric BCE loss is used for pixel-level mask supervision; $\mathcal{L}_{BCE}$ is the default BCE loss for class prediction; and $\lambda_{box}, \lambda_{mask}, \lambda_{cls} \in \mathbb{R}^+$ are weighting coefficients for each term.

The asymmetric mask loss is specifically designed to handle occlusion by penalising false negatives more heavily than false positives, encouraging the model to preserve partial object masks. It is defined as:

$$\mathcal{L}_{asym-BCE}(p, y) = -[\alpha_{FN} \cdot y \cdot \log(p) + \alpha_{FP} \cdot (1-y) \cdot \log(1-p)] \quad (6)$$

where $p \in [0,1]$ is the predicted probability; $y \in [0,1]$ is the ground truth label for each pixel; $\alpha_{FN}$ is the asymmetry coefficient for the positive (foreground) class, empirically set to 1.1; and $\alpha_{FP}$ is for the negative (background) class, empirically set to 0.9. By integrating this asymmetric formulation, the model is biased toward generating more complete masks under occlusion, ensuring segmentation accuracy in dense and cluttered agricultural environments.

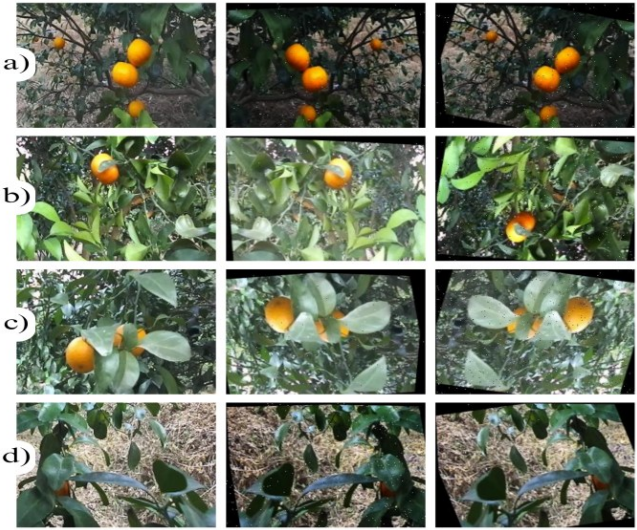

Fig. 3. Data augmentation examples of occlusion sensitive subsets of a) zero-level occlusion, b) low-level occlusion, c) medium-level occlusion and d) high-level occlusion.

## III. DATA AND TRAINING STRATEGY

### A. Dataset preparation

This study employs a dataset comprising 1,000 RGB images captured in an outdoor citrus orchard, naturally exhibiting varying degrees of occlusion. The images originate from a publicly accessible dataset designed by Hou et al. [27] for citrus detection tasks. A total of 3,749 citruses were manually annotated using polygon-based instance segmentation masks via the Roboflow platform. To support amodal segmentation, each annotation was deliberately extended to include the full extent of the fruit, even when parts were visually occluded. The dataset was subsequently partitioned into 700 images for training, 200 for validation, and 100 for testing. No data augmentation techniques or post-processing steps were applied for this partition. This dataset will be referred to as the initial dataset throughout the remainder of this paper.

To prepare occlusion-sensitive subsets, 71 images with a 256×256-pixel size were manually cropped from the 100 test images by selecting the fruit regions. The cropped images were grouped by occlusion level and annotated with previously described amodal masks. Unlike the initial dataset annotations, this set was augmented using horizontal and vertical flips, rotations between ±15°, shear of ±10°, exposure variation between −15% and +15%, and random noise affecting up to 1.45% of pixels. This process tripled the image count for each occlusion class. The final occlusion-sensitive subset included 42 images with zero occlusion (fully visible), 96 with low occlusion (∼0-20%), 51 with medium occlusion (∼20-50%), and 45 with high occlusion (∼over 50%). An example of this preparation is illustrated in Fig. 3.

### B. Implementation Details

The experiments were conducted using Python 3.8.20 and PyTorch 2.4.1 with CUDA 12.1 support. Training and validation were performed on a system featuring an NVIDIA GeForce RTX 3060 GPU with 12 GB of VRAM. The model was trained for 100 epochs with a batch size of 8. Stochastic Gradient Descent (SGD) was used as the optimiser, configured with an initial learning rate of 0.01, a momentum of 0.937, and a weight decay of 0.0005. The loss weight coefficients for bounding box regression, mask segmentation, and classification were maintained at their default values, consistent with standard YOLO11n training configurations. No transfer learning was applied, and the model was trained from scratch using the initial dataset.

### C. Model Evaluation Metrics

This study evaluates the proposed model's performance using Precision (P), Recall (R), mean Average Precision at IoU threshold of 50% (mAP@50), and mean Average Precision calculated over IoU thresholds ranging from 50% to 95% (mAP@50:95). The corresponding performance metrics are as described in Eqs. (7–10),

$$P = \frac{TP}{TP + FP} \quad (7)$$

$$R = \frac{TP}{TP + FN} \quad (8)$$

$$AP = \int_0^1 P(R) dR \quad (9)$$

$$mAP = \frac{1}{n} \sum_{i=1}^{n} AP_i \quad (10)$$

where True Positives (TP) represent correctly identified positive instances; False Positives (FP) correspond to negative instances incorrectly classified as positive; and False Negatives (FN) indicate positive instances missed by the model. Average Precision (AP) is calculated as the area under the Precision–Recall curve for each class, while mean Average Precision (mAP) is the average of AP values across all classes; and $i = 1,2,3,...,n$ which $n$ represents the class number. Additionally, inference time measured in milliseconds was included to assess the real-time applicability of the model. The model's complexity and size were further characterised by the number of floating-point operations (GFLOPs) and the total parameter count (in millions(M)), respectively, providing insights into computational efficiency and scalability.

## IV. FRUIT HARVESTING

To verify the effectiveness of the proposed model under real-world-inspired conditions, harvesting experiments were conducted across varying levels of occlusion. The tests were performed in a controlled laboratory environment to ensure consistency and repeatability. As shown in Fig. 4a, the experimental setup included an Emika Franka Panda robotic arm equipped with an Intel RealSense D415 depth camera, interacting with an artificial tree to simulate robotic picking tasks. To strictly evaluate the perception system's performance, the mechanical complexity of fruit detachment was simplified. Real fruits were mounted on the outer periphery of the synthetic tree using low-resistance metal fixtures, as explicitly depicted in Fig. 4a. Fig. 4b (1–4) illustrates a group of real fruit examples that were used in the picking experiments, classified into zero (fully visible (Fig. 4b (1)), low (Fig. 4b (2)), medium (Fig. 4b (3)), and high (Fig. 4b (4)) occlusion levels, reflecting increasing visual complexity. This setup physically replicates the occlusion-sensitive subsets described in Section III, using the same occlusion ratio ranges. The complete target fruit localisation methodology is summarised in Fig. 4c. In the harvesting experiments, this categorisation enabled a structured comparison between segmentation performance and picking success across each occlusion level. For each occlusion class, 54 fruits were presented, resulting in 216 trials per model and a total of 432.

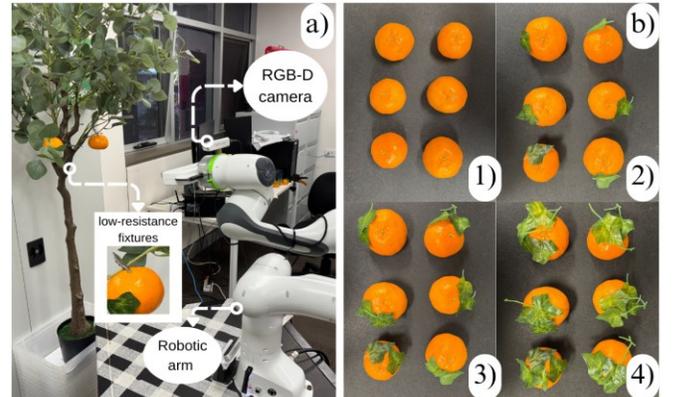

Fig. 4. Harvesting experiments setup; a) Artificial tree and Emika Franka Panda cobot, b) real fruits used in the picking experiments, categorised into 1) zero (fully visible, 2), low, 3) medium, and 4) high occlusion levels.

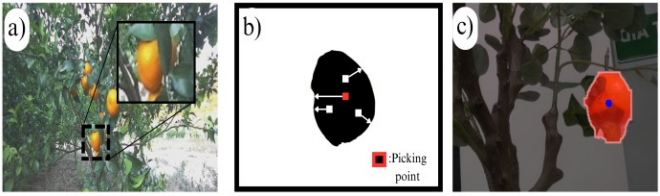

Fig. 5. Picking point identification: a) occluded fruit instance in dataset b) Distance transform-based picking point identification process and, c) amodal mask generation with defined picking point.

To physically validate the GDA-YOLO11's segmentation performance across increasing occlusion exposure on occlusion-sensitive subsets and to quantify the overall outcome of the harvesting pipeline, the harvesting success metric H was defined as shown in Eq. (11).

$$H = \frac{N_{picked}}{N_{total}} \quad (11)$$

where $N_{picked}$ denotes the number of successfully picked citrus; and $N_{total}$ represents the total number of citrus that have been included in the harvesting experiments.

*B. Picking point identification*

To estimate the accurate picking point of the partially occluded target fruit, a Euclidean distance transform-based method was employed [28]. Formalising this approach, the optimal picking point $P^*$ is determined by solving the following maximisation problem:

$$P^* = \arg\max_{(x,y)\in\mathcal{M}} (\|(x,y) - \mathcal{B}\|_2) \quad (12)$$

where $\mathcal{M}$ represents the set of foreground pixels constituting the binary instance mask, $\mathcal{B}$ denotes the set of background pixels, and $\|\cdot\|_2$ represents the Euclidean norm. This method prioritises the most stable region of the visible fruit segment by maximising clearance from object boundaries. Fig. 5 explains the overall process: an amodal instance segmentation mask has been generated from an occluded fruit instance (Fig. 5a), and the picking point has been identified based on Eq. (12) as illustrated in Fig. 5b. Finally, at the end of the process, the defined picking point in the real application obtained from the RGB-D camera frame can be seen in Fig. 5c. This approach offers greater stability than moment-based centroids, which may be distorted when objects are partially hidden.

*C. Target Fruit Localisation and Manipulation*

In the robotic setup, an RGB-D camera is mounted near the end of the robotic arm in an eye-in-hand configuration, meaning the camera moves jointly with the arm during the harvesting operation. This configuration allows positional transformation directly from the robot, using its known joint positions provided through the manufacturer's official library, libfranka.

The localisation process begins with the segmentation of the target fruit in the image frame. The fruit's picking point is determined, along with its associated depth value, from the RGB-D camera. Using this information, the 3D coordinates of the fruit in the camera coordinate frame are obtained via back-projection using the image projection model:

$$\begin{bmatrix} u \\ v \\ w \end{bmatrix} = \begin{bmatrix} \mathbf{K} & \mathbf{0} \\ \mathbf{0} & 1 \end{bmatrix} \begin{bmatrix} \mathbf{R} & \mathbf{t} \\ \mathbf{0} & 1 \end{bmatrix} \begin{bmatrix} X \\ Y \\ Z \\ 1 \end{bmatrix} \quad (13)$$

where $u$ and $v$ are the pixel coordinates of the fruit in the image, and $w$ is a homogeneous scale factor related to the fruit's depth. The vector $[X, Y, Z]^T$ represents the fruit's position in the world coordinate frame. The intrinsic matrix $\mathbf{K}$ contains the focal lengths and principal point of the camera, and the extrinsic transformation $[\mathbf{R}|\mathbf{t}]$ represents the rotation and translation between coordinate frames. The intrinsic parameters were determined through intrinsic camera calibration, while the extrinsic transformation was obtained through hand–eye calibration, which accounts for the fixed offset between the camera and the end-effector.

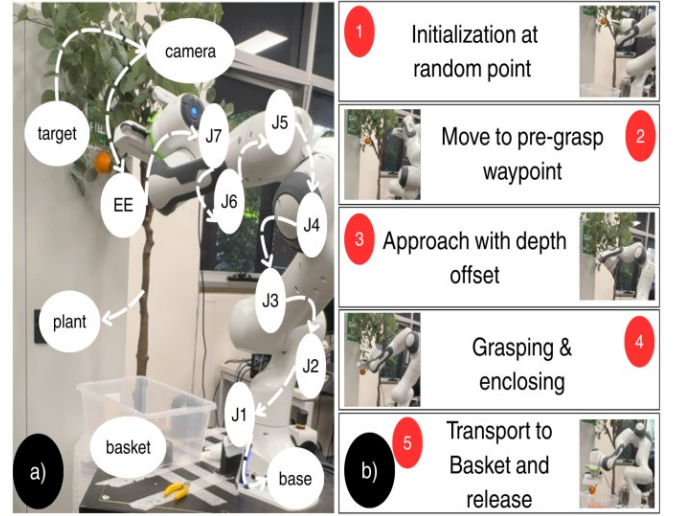

Fig. 6. Visualisation of the proposed robotic localisation and manipulation framework. (a) The coordinate transformation chain for target localisation, illustrating the mapping from 2D image coordinates to the 3D robot base frame via intrinsic back-projection and kinematic chain transformations. (b) The sequential execution of the harvesting strategy: (1) Initialisation at home configuration, (2) Alignment with the pre-grasp waypoint under fixed orientation, (3) Linear approach with a +2 cm safety offset to ensure full enclosure, (4) Grasping actuation, and (5) Transport to the collection basket and release.

Once the 3D position is calculated in the camera frame, it is first transformed into the end-effector frame using the calibrated hand–eye transformation matrix. This accounts for the physical displacement between the RGB-D sensor and the robot's end-effector. Subsequently, the fruit's position is transformed into the robot base frame by following the full kinematic chain of the robotic arm, as shown in Fig.6.

Once the 3D position is calculated in the camera frame, it is first transformed into the end-effector frame using the calibrated hand–eye transformation matrix. This accounts for the physical displacement between the RGB-D sensor and the robot's end-effector. Subsequently, the fruit's position is transformed into the robot base frame by following the full kinematic chain of the robotic arm, as shown in Fig.6.

To ensure a robust grasp, the approach strategy was decomposed into a two-stage motion primitive. First, the end-effector moves to a pre-grasp waypoint aligned with the target's centroid but offset by a safety margin. From this waypoint, a linear approach vector is executed to reach the final grasping coordinates. A fixed end-effector orientation ($\{-\pi, -\frac{\pi}{2}, 0\}$) relative to the base frame was maintained during this phase, simplifying the path planning by reducing the search space to 3D Cartesian position control.

The motion planning for the harvesting operation was executed using Cartesian-space control with quintic polynomial velocity profiles to ensure smooth trajectories. The end-effector moved from the home position to the computed target coordinates using a deterministic point-to-point motion generator (via LibOrl [29]). In this experimental setup, explicit dynamic collision avoidance algorithms were not employed. Instead, as an additional safety measure, target fruits were positioned on the outer periphery of the synthetic tree to mitigate manipulation and grasping difficulties arising from occlusion, ensuring an obstacle-free workspace for the pre-planned Cartesian paths. Finally, to secure the fruit, a +2 cm forward position offset was applied along the approach axis, ensuring the gripper fingers fully enclosed the target before actuation. Consequently, robotic action focused on 'approach and enclose' rather than complex grasping strategies, preventing mechanical compliance from compensating for localisation errors. Although the robotic framework supports manipulation primitives such as rotation, these were disabled for this experiment to isolate the grasping success as a direct function of localisation accuracy, rather than mechanical gripping force or stem resistance.

## V. RESULTS AND DISCUSSION

### A. GDA-YOLO11

The experimental results demonstrate the effectiveness of the proposed improvements on the model's performance. GDA-YOLO11 achieved 84.4% precision, 84.6% recall, 91.4% mAP@50, and 63.6% mAP@50:95 on the initial validation set, demonstrating strong overall performance in the amodal instance segmentation task. As shown in Table I, the proposed updates systematically improve the baseline model. Incorporating GAM, SPPF extension, deep head, and asymmetric mask loss leads to an over 5.1% increase in precision and a 1.3% rise in mAP@50. Notably, the use of double GAM contributed to approximately a 1% boost in the most sensitive metric, mAP@50:95, due to their ability to capture global contextual information and enhance feature representation under occluded conditions. Although the recall remains comparable to YOLO11's, it is marginally lower; however, the ablation experiment results indicate the loss function contributes a positive effect of around 0.8% on the recall score. While GFLOPs increased due to the added extra complexity of the deep head, the number of the model's parameters increased by approximately 18%. Despite these modifications, the model maintains real-time inference speed with only a minimal 1.3 ms increase in total inference time. This increase primarily originates from the deepening of the segmentation head, which introduces additional parameters and slightly raises the GFLOPs count. At the same time, this structural enhancement is also the key factor behind the improvement observed in the mAP@50 and mAP@50:95 metrics. These findings confirm that the enhancements

TABLE I
ABLATION EXPERIMENTS[1]

| M | GAM | S | DH | L | C | P | R | @50 | @50:95 | T | G | PN |
|---|---|---|---|---|---|---|---|---|---|---|---|---|
| B | × | × | × | × | ✓ | 0.793 | **0.853** | 0.901 | 0.627 | 5.5 | 10.2 | 2.83 |
| G-v1 | ✓ | × | × | × | ✓ | 0.824 | 0.842 | 0.906 | 0.628 | 4.7 | 10.2 | 2.83 |
| G-v2 | ✓✓ | × | × | × | × | 0.839 | 0.825 | 0.904 | **0.636** | 5.5 | 10 | 2.60 |
| G-v3 | ✓✓ | ✓ | × | × | × | 0.833 | 0.839 | 0.901 | 0.63 | 5.3 | 10 | 2.60 |
| GD-v1 | ✓✓ | × | ✓ | × | × | 0.815 | 0.841 | 0.907 | 0.632 | 7 | 20.5 | 3.34 |
| GD-v2 | ✓✓ | ✓ | ✓ | × | × | 0.834 | 0.834 | 0.903 | 0.632 | 7 | 20.5 | 3.34 |
| GDA | ✓✓ | ✓ | ✓ | ✓ | × | **0.844** | 0.846 | **0.914** | **0.636** | 6.8 | 20.5 | 3.34 |

TABLE II
COMPARISON OF OTHER OCCLUSION-AWARE MODELS[2]

| Model | P | R | @50 | @50:95 | T | G | PN |
|---|---|---|---|---|---|---|---|
| [30] | 0.817 | 0.816 | 0.889 | 0.607 | 5.2 | 12 | 3.25 |
| [31] | 0.816 | 0.841 | 0.903 | 0.629 | 5.8 | 7.6 | 4.44 |
| [32] | 0.791 | 0.845 | 0.902 | 0.633 | 4.8 | 12 | 3.25 |
| [33] | 0.816 | 0.824 | 0.896 | 0.61 | 5.1 | 16 | 2.96 |
| [34](s) | 0.814 | **0.856** | **0.915** | **0.654** | 8.1 | 35.3 | 10.1 |
| [34](n) | 0.793 | 0.853 | 0.901 | 0.627 | 5.5 | 10.2 | 2.83 |
| (ours) | **0.844** | 0.846 | 0.914 | 0.636 | 6.8 | 20.5 | 3.34 |

## TABLE III
### HARVESTING EXPERIMENT RESULTS[3]

| M | O | @50 | @50:95 | H | H (%) |
|---|---|---|---|---|---|
| YOLO11 | Z | 0.886 | 0.543 | 52 | 96.29 |
| YOLO11 | L | 0.88 | 0.526 | 46 | 85.18 |
| YOLO11 | M | 0.564 | 0.218 | 24 | 44.44 |
| YOLO11 | H | 0.341 | 0.112 | 10 | 18.51 |
| GDA - YOLO11 | Z | 0.872 | 0.537 | 50 | 92.59 |
| GDA - YOLO11 | L | 0.888 | 0.522 | 46 | 85.18 |
| GDA - YOLO11 | M | 0.569 | 0.25 | 26 | 48.14 |
| GDA - YOLO11 | H | 0.372 | 0.119 | 12 | 22.22 |

[1] '×' indicates that the corresponding study does not implement the process; '√' signifies that the process is implemented; '√√' denotes that the process is applied in two distinct stages within the study; bold values indicate the best results in each category. "M" is the model name. "GAM" means the "GAM" module, "S" is the "SFPP" block, and "DH" is the Deep head. "L" stands for the Asymmetric mask loss function, and "C" is "C2PSA". Finally, "B" is the "YOLOv11" base model, and terms like "G-v", "GD-v", and "GDA" refer to the different model versions developed in this work. [1, 2] For performance metrics, "P" is Precision and "R" is Recall. "@50" means the mAP score at 0.50 IoU, while "@50:95" is the mAP score across different IoU levels. For computational cost, "T" is time, "G" is GFLOPs, and "PN" is the number of parameters. [3] O' represents the occlusion classes, denoted as "Z", "L", "M", and "H" for zero, low, medium, and high levels, respectively. 'H' shows the harvesting success out of 54 trials, while 'H (%)' is the corresponding percentage.

effectively address occlusion challenges without sacrificing overall efficiency or speed.

To further validate the model's effectiveness, GDA-YOLO11 was compared against several state-of-the-art nano-scale(n) and small-scale(s) architectures trained on the initial dataset and under the same experimental conditions, as shown in Table II. GDA-YOLO11 achieved the highest precision score with 0.844, outperforming models like NVW-YOLOv8n with 0.817, SGW-YOLOv8n with 0.816, and even the larger YOLO11s model, which reached 0.814. GDA-YOLO11 remained highly competitive with a recall of 0.846, suppressing all custom models as well as the YOLOv8n in recall performance. Also, it achieved a mAP@50 of 0.914, surpassing all nano-scale alternatives. Additionally, the proposed model nearly matched the 0.915 score of YOLO11s while using significantly fewer parameters, with 3.34 million and offering faster inference at 6.8 milliseconds compared to YOLO11s' 8.1 milliseconds. Additionally, GDA-YOLO11 recorded the highest mAP@50:95 among all n-scale models, with a score of 0.636. These results confirm that GDA-YOLO11 not only outperforms comparable lightweight models across all major performance indicators but also approaches the accuracy of larger architectures without compromising speed or efficiency. This balance makes it highly suitable for practical deployment in agricultural harvesting scenarios where real-time inference and compact model size are critical.

*B. Harvesting experiments*

During the harvesting experiments, artificially occluded citrus fruits were picked using the proposed framework. Under zero occlusion, YOLO11 achieved 52 successful picks (96.29% H), slightly outperforming GDA-YOLO11, which resulted in 50 successful picks (92.59% H). However, in low occlusion, both models reached identical harvesting performance, completing 46 picks with an 85.18% success rate, indicating that minor foliage does not significantly differentiate the models. As occlusion severity increased, GDA-YOLO11 demonstrated more robust performance. Under medium occlusion, it attained 26 successful picks (48.14% H) versus YOLO11's 24 picks (44.44% H). In high occlusion, where the task is most challenging, GDA-YOLO11 completed 12 successful picks (22.22%) compared to YOLO11's 10 successful picking attempts (18.51%). While the absolute numerical margin appears modest, it represents a substantial relative improvement in robustness when visual cues are scarce.

To correctly interpret these results, it is essential to define the failure modes. Crucially, due to the low-resistance experimental setup described in Section IV-A, zero mechanical interaction failures (e.g., gripper slip or insufficient force) were observed; thus, all failures were attributable to perception limitations. Detailed inspection revealed that, in high occlusion trials, only three failures were due to inaccurate or low-quality masks (localisation errors) where the gripper missed the fruit volume. The remaining failures occurred because no valid mask could be generated due to extreme occlusion (detection failures). In contrast, no such failures were observed in the zero and low occlusion groups, where all masks were successfully generated. Instead, the few unsuccessful cases in these groups were explicitly identified as localisation errors stemming from suboptimal mask quality, which led to the computation of inaccurate picking points under varying fruit orientations.

Finally, the experiments revealed a strong positive correlation ($R^2 \approx 0.986$) between the segmentation accuracy (mAP@50) and the physical harvesting success rate (H). This finding aligns with trends reported in the literature ($R^2 \approx 0.940$) [17] and validates the hypothesis that segmentation quality is a reliable proxy for robotic picking performance in controlled environments. Furthermore, the sharp decline in success rates beyond 50% occlusion (medium and high levels) reinforces prior findings [35] that this occlusion ratio represents a critical harvestability threshold for current vision systems.

## VI. CONCLUSION

This study presents a robotic harvesting framework based on amodal instance segmentation for occlusion handling, which is one of the most critical tasks in vision-driven robotic harvesting. A novel amodal instance segmentation model was introduced and evaluated specifically under varying occlusion conditions using both segmentation benchmarks and realistic harvesting experiments. In this regard, the study bridges amodal perception and robotic action by generating instance-level amodal masks through a dedicated deep learning without additional geometric processing or reconstruction steps. This unified framework enables a perception-driven and occlusion-robust harvesting and represents, to the best of our knowledge, the first real-world validation of an amodal segmentation approach for fruit harvesting since its initial application to fruit imagery in 2023. By eliminating intermediate steps such as 3D reconstruction, shape fitting or mask refinement, the proposed method simplifies the system architecture and preserves robustness against occlusion, making it more scalable and deployable in complex agricultural environments. While this work demonstrates the feasibility of amodal perception for robotic harvesting, further studies are needed to improve model performance, particularly under high-level occlusion and to explore more advanced amodal segmentation strategies. Additionally, RGB-D cameras, despite their widespread use and advantages, remain insufficient for handling full occlusion scenarios, indicating the need for alternative sensing or reasoning mechanisms in future work.

## REFERENCES

[1] United Nations, "Key findings," 2019. [Online]. Available: https://population.un.org/wpp/Publications/Files/WPP2019_10KeyFindings.pdf. [Accessed: Apr. 19, 2025].
[2] R. Ishangulyyev, S. Kim, and S. H. Lee, "Understanding Food Loss and Waste—why Are We Losing and Wasting food?," Foods, vol. 8, no. 8, p. 297, Jul. 2019.
[3] F. Economou et al., "The concept of food waste and food loss prevention and measuring tools," Waste Manag. Res., vol. 42, no. 8, Mar. 2024.
[4] WWF, "DRIVEN TO WASTE: THE GLOBAL IMPACT OF FOOD LOSS AND WASTE ON FARMS," 2021. [Online]. Available: https://wwfint.awsassets.panda.org/downloads/wwf_uk__driven_to_waste___the_global_impact_of_food_loss_and_waste_on_farms.pdf
[5] X. Yu et al., "A lab-customized autonomous humanoid apple harvesting robot," Comput. Elect. Eng, vol. 96, p. 107459, Dec. 2021.
[6] L. Ma et al., "A Method of Grasping Detection for Kiwifruit Harvesting Robot Based on Deep Learning," Agronomy, vol. 12, no. 12, pp. 3096–3096, Dec. 2022.
[7] Y. Qian, R. Jiacheng, W. Pengbo, Y. Zhan and G. Changxing, "Real-time detection and localization using SSD method for oyster mushroom picking robot," in 2020 IEEE International Conference on Real-time Computing and Robotics (RCAR), Asahikawa, Japan, 2020, pp. 158-163.
[8] A. Stepanova et al., "Harvesting tomatoes with a Robot: an evaluation of Computer-Vision capabilities," in 2023 IEEE International Conference on Autonomous Robot Systems and Competitions (ICARSC), Tomar, Portugal, 2023, pp. 63-68.
[9] H. Sun, B. Wang, and J. Xue, "YOLO-P: An efficient method for pear fast detection in complex orchard picking environment," Front. Plant Sci., vol. 13, Jan. 2023.
[10] Y. Bai, Y. Guo, Q. Zhang, B. Cao, and B. Zhang, "Multi-network fusion algorithm with transfer learning for green cucumber segmentation and recognition under complex natural environment," Comput. Electron. Agric., vol. 194, pp. 106789–106789, Mar. 2022.
[11] C. Chen, B. Li, J. Liu, T. Bao, and N. Ren, "Monocular positioning of sweet peppers: An instance segmentation approach for harvest robots," Biosyst. Eng., vol. 196, pp. 15–28, Aug. 2020.
[12] J. Wei et al., "Tomato ripeness detection and fruit segmentation based on instance segmentation," Front. Plant Sci., vol. 16, p. 1503256, May 2025.
[13] Z. Zhao, Y. Hicks, X. Sun, and C. Luo, "FruitQuery: A lightweight query-based instance segmentation model for in-field fruit ripeness determination," Smart Agric. Technol., vol. 12, p. 101068, Jun. 2025.
[14] S. Kim, S.-J. Hong, J. Ryu, E. Kim, C.-H. Lee, and G. Kim, "Application of amodal segmentation on cucumber segmentation and occlusion recovery," Comput. Electron. Agric., vol. 210, pp. 107847–107847, Jul. 2023.
[15] T. Yu, C. Hu, Y. Xie, J. Liu, and P. Li, "Mature pomegranate fruit detection and location combining improved F-PointNet with 3D point cloud clustering in orchard," Comput. Electron. Agric., vol. 200, p. 107233, Sep. 2022.
[16] X. Wang, H. Kang, H. Zhou, W. Au, M. Y. Wang, and C. Chen, "Development and evaluation of a robust soft robotic gripper for apple harvesting," Computers and Electronics in Agriculture, vol. 204, p. 107552, 2023.
[17] L. Gong, W. Wang, T. Wang, and C. Liu, "Robotic harvesting of the occluded fruits with a precise shape and position reconstruction approach," Journal of Field Robotics, vol. 39, no. 1, pp. 69–84, Oct. 2021.
[18] J. Liang, K. Huang, H. Lei, Z. Zhong, Y. Cai, and Z. Jiao, "Occlusion-aware fruit segmentation in complex natural environments under shape prior," Computers and Electronics in Agriculture, vol. 217, p. 108620, Jan. 2024.
[19] E. Kok and C. Chen, "Occluded apples orientation estimator based on deep learning model for robotic harvesting," Computers and Electronics in Agriculture, vol. 219, p. 108781, Apr. 2024.
[20] F. Magistri et al., "Contrastive 3D Shape Completion and Reconstruction for Agricultural Robots Using RGB-D Frames," IEEE Robotics and Automation Letters, vol. 7, no. 4, pp. 10120–10127, Jul. 2022.
[21] T. Li, Q. Feng, Q. Qiu, F. Xie, and C. Zhao, "Occluded Apple Fruit Detection and Localization with a Frustum-Based Point-Cloud-Processing Approach for Robotic Harvesting," Remote Sensing, vol. 14, no. 3, p. 482, Jan. 2022.
[22] P. M. Blok, E. J. van Henten, F. K. van Evert, and G. Kootstra, "Image-based size estimation of broccoli heads under varying degrees of occlusion," Biosyst. Eng., vol. 208, pp. 213–233, Aug. 2021.
[23] J. Gené-Mola et al., "Looking behind occlusions: A study on amodal segmentation for robust on-tree apple fruit size estimation," Comput. Electron. Agric., vol. 209, p. 107854, Jun. 2023.
[24] J. Yang, H. Deng, Y. Zhang, Y. Zhou, and T. Miao, "Application of amodal segmentation for shape reconstruction and occlusion recovery in occluded tomatoes," Front. Plant Sci., vol. 15, p. 1376138, Jun. 2024
[25] J. Ao, Q. Ke, and K. A. Ehinger, "Amodal Intra-class Instance Segmentation: Synthetic Datasets and Benchmark," 2022 IEEE/CVF Winter Conference on Applications of Computer Vision (WACV), pp. 280–289, Jan. 2024.
[26] Y. Liu, Z. Shao, and N. Hoffmann, "Global Attention Mechanism: Retain Information to Enhance Channel-Spatial Interactions," arXiv.org, Dec. 10, 2021. [Online]. https://arxiv.org/abs/2112.05561.
[27] C. Hou et al., "Detection and localization of citrus fruit based on improved You Only Look Once v5s and binocular vision in the orchard," Front. Plant Sci., vol. 13, p. 972445, Jul. 2022.
[28] G. Borgefors, "Distance transformations in arbitrary dimensions," Comput. Gr. Image Process., vol. 27, no. 3, pp. 321–345, Sep. 1984.
[29] M. Boneberger, M. von Unwerth, and P. Mohammadi, liborl (GitHub repository), 2020. [Online]. Available: https://github.com/marcbone/liborl. [Accessed: Apr. 17, 2025].
[30] A. Wang et al., "NVW-YOLOv8s: An improved YOLOv8s network for real-time detection and segmentation of tomato fruits at different ripeness stages," Comput. Electron. Agric., vol. 219, pp. 108833–108833, Apr. 2024.
[31] T. Wu, Z. Miao, W. Huang, W. Han, Z. Guo, and T. Li, "SGW-YOLOv8n: An Improved YOLOv8n-Based Model for Apple Detection and Segmentation in Complex Orchard Environments," Agriculture, vol. 14, no. 11, pp. 1958–1958, Oct. 2024.
[32] G. Jocher, A. Chaurasia, and J. Qiu, "YOLOv8 by Ultralytics," GitHub, 2023. [Online]. Available: https://github.com/ultralytics/ultralytics. [Accessed: May 14, 2025].
[33] Z. Tian, H. Hao, G. Dai, and Y. Li, "Optimizing tomato detection and counting in smart greenhouses: A lightweight YOLOv8 model incorporating high- and low-frequency feature transformer structures," Netw. Comput. Neural Syst., pp. 1–37, Nov. 2024.
[34] G. Jocher and J. Qiu, "Ultralytics YOLO11," GitHub, software ver. 11.0.0, 2024. [Online]. Available: https://github.com/ultralytics/ultralytics [Accessed: May 14, 2025].
[35] J. Hemming, J. Ruizendaal, J. Hofstee, and E. van Henten, "Fruit Detectability Analysis for Different Camera Positions in Sweet-Pepper," Sensors, vol. 14, no. 4, pp. 6032–6044, Mar. 2014.